\documentclass[conference]{IEEEtran}
\IEEEoverridecommandlockouts
\usepackage{cite}
\usepackage{amsmath,amssymb,amsfonts}
\usepackage{algorithmic}
\usepackage{graphicx}
\usepackage{textcomp}
\usepackage{xcolor}
\usepackage{url} 
\usepackage{booktabs}
\usepackage{subcaption}
\usepackage[normalem]{ulem}
\def\BibTeX{{\rm B\kern-.05em{\sc i\kern-.025em b}\kern-.08em
    T\kern-.1667em\lower.7ex\hbox{E}\kern-.125emX}}
\begin{document}

\title{CoVeRaP: Cooperative Vehicular Perception through mmWave FMCW Radars}

\author{
Jinyue Song$^{\star}$ \qquad
Hansol Ku$^{\star}$ \qquad 
Jayneel Vora$^{\diamond}$ \qquad  
Nelson Lee$^{\star}$ \qquad \\
Ahmad Kamari$^{\ddagger}$ \qquad  
Prasant Mohapatra$^{\dagger}$ \qquad 
Parth Pathak$^{\ddagger}$ 
 \\
    \small
    $^{\star}$ UC Davis, CA, USA, \{jysong,solku,nplee\}@ucdavis.edu\\
    $^{\diamond}$ Intel Corporation, CA, USA, jayneelvora98@gmail.com\\
    $^{\dagger}$ USF Tampa, FL, USA, pmohapatra@usf.edu\\
    $^{\ddagger}$ GMU, VA, USA, \{akamari,phpathak\}@gmu.edu
}

\maketitle

\begin{abstract}

Automotive FMCW radars remain reliable in rain and glare, yet their sparse, noisy point clouds constrain 3-D object detection. We therefore release CoVeRaP, a 21 k-frame cooperative dataset that time-aligns radar, camera, and GPS streams from multiple vehicles across diverse manoeuvres. Built on this data, we propose a unified cooperative-perception framework with middle- and late-fusion options. Its baseline network employs a multi-branch PointNet-style encoder enhanced with self-attention to fuse spatial, Doppler, and intensity cues into a common latent space, which a decoder converts into 3-D bounding boxes and per-point depth confidence. Experiments show that middle fusion with intensity encoding boosts mean Average Precision by up to 9× at IoU 0.9 and consistently outperforms single-vehicle baselines. CoVeRaP thus establishes the first reproducible benchmark for multi-vehicle FMCW-radar perception and demonstrates that affordable radar sharing markedly improves detection robustness. Dataset and code are publicly available to encourage further research.

\end{abstract}


\section{Introduction}

Recent advances in automotive technology have enabled sophisticated driver assistance and autonomous driving systems \cite{adaptive, lane, blind}. Companies such as Waymo and Tesla have propelled these innovations forward, yet robust perception in complex, real-world environments remains a critical challenge. While traditional sensors like cameras and LiDAR offer high-resolution data, they can fail under adverse conditions. Frequency-Modulated Continuous Wave (FMCW) radars, in contrast, deliver reliable measurements in diverse weather and lighting conditions; however, single-radar systems often yield sparse, noisy point clouds and provide partial views of objects.

Cooperative perception has emerged as a promising solution, particularly in a Vehicle-to-Vehicle (V2V) context, where sharing radar data from multiple vehicles can enhance environmental understanding and reduce blind spots. Fusing data from different viewpoints can generate denser point clouds and yield more accurate 3D bounding-box detections, thereby improving object tracking and overall situational awareness.

In complex traffic scenarios, single-sensor modalities often fall short due to occlusions and dynamic conditions. Cooperative perception addresses these challenges by sharing radar data among vehicles, enhancing situational awareness and reducing blind spots. The fused data yield richer point clouds and more reliable 3D bounding-box detections, thereby boosting safety and system robustness. Furthermore, prior work~\cite{aditya} on memory-efficient 3D segmentation underscores the importance of lightweight architectures for real-time applications, aligning with our approach.

Despite its promise, cooperative perception faces several challenges. Radar data are normally sparse and noisy, often capturing only fragments of objects. Robust sensor synchronization—both within and across vehicles—is essential, yet differences in sensor clock rates, spatial positions, and calibration errors complicate this task. Constructing a multi-vehicle dataset further demands precise data alignment of heterogeneous sensors, effective management of temporal drift, and extensive data cleaning with accurate ground truth annotations. In addition, various data fusion strategies—whether early, middle, or late fusion—each bring distinct trade-offs between feature integration and model performance.

Our work makes three key contributions: 
\begin{itemize} 
\item \textbf{CoVeRaP Dataset:} We introduce a novel large-scale cooperative radar perception dataset that integrates synchronized multi-vehicle radar data with camera images and GPS information, featuring high-quality ground truth annotations and structured data splits. 
\item \textbf{Unified System Architecture for Cooperative Sensing:} We propose an end-to-end pipeline that captures, synchronizes, and fuses sensor data from multiple vehicles. Our architecture supports both feature-level (middle fusion) and prediction-level (late fusion) strategies. 
\item \textbf{Baseline 3D Bounding-Box Detection Model and Evaluation:} We present a baseline model based on a multi-branch architecture with attention mechanisms. Our experiments demonstrate that cooperative fusion, particularly through middle fusion, substantially improves detection performance compared to single-view approaches. 
\end{itemize}


The remainder of this paper is organized as follows. Section~\ref{sec:background} and Section~\ref{sec:related_work} review the necessary background and related work in cooperative perception and radar-based sensing. Section~\ref{sec:dataset} details the CoVeRaP dataset, including data collection, sensor synchronization, and dataset construction. In Section~\ref{sec:system_and_model}, we describe our system architecture and baseline 3D bounding-box detection model, with an emphasis on fusion strategies. Section~\ref{sec:experiment_and_results} presents our experimental evaluation, and finally, Section~\ref{sec:conclusion} concludes with a summary of our findings, limitations, and directions for future research.
\section{Background}
\label{sec:background}

This section reviews key concepts in FMCW-radar cooperative perception and summarizes popular fusion strategies, setting the stage for our dataset and baseline model.

\subsection{Fundamentals of FMCW Radars and 3D Point Clouds}




Frequency-Modulated Continuous Wave (FMCW) radars\cite{instruments2020fundamentals} measure distance, velocity, and angular position by transmitting a “chirp” whose frequency increases linearly over a brief time interval. When this chirp reflects off a target and returns, the round-trip time introduces a delay, producing an intermediate frequency (IF) signal. By applying Fast Fourier Transforms (FFT) to the IF signal, the radar determines the range (via a range FFT), velocity (via a Doppler FFT), and angular position (via an angle FFT). These outputs collectively form a 3D point cloud, which is vital for tasks such as object tracking and scene understanding in autonomous systems.

\subsection{Fusion Strategies in Cooperative Perception}
Effective data fusion is critical in cooperative perception. We broadly distinguish three fusion strategies:

Early Fusion: Raw sensor data from multiple vehicles are combined before feature extraction. Although this retains maximum information, it is challenged by large data volumes and the need for high-bandwidth communication, as well as difficulties in synchronizing raw radar waveforms.

Middle Fusion: Each vehicle processes its own sensor data to extract features (e.g., point clouds, range, velocity, signal intensity), and these features are then merged in a common spatial frame. Middle fusion strikes a balance between preserving rich information and addressing practical constraints, though it requires precise synchronization and alignment.

Late Fusion: Initially, separate models are trained on each vehicle’s sensor data, then the final predictions (e.g., 3D bounding boxes) are combined. This strategy is more robust to sensor noise and discrepancies but relies on techniques such as confidence weighting or voting to reconcile differences.

\subsection{Point-based Models for 3D Point Cloud}
Point-based deep learning models operate directly on raw 3D points without voxelization or image transformations. A seminal example is PointNet~\cite{qi2017pointnet}, which introduced a permutation-invariant architecture that processes unordered point sets, revealing that raw point data alone can support effective feature extraction. Building on this foundation, subsequent methods refine local neighborhood analysis and exploit richer geometric cues, making point-based models fundamental tools in 3D perception.

For radar perception, point-based models excel because they skip voxelization, preserving fine geometric details while lowering computation. Their ability to adapt to varying point densities matches the fluctuating radar returns across scenes, enabling stronger feature extraction and more reliable object detection in challenging conditions.

\section{Related Work}
\label{sec:related_work}
\subsection{Deep Point-based Approaches for 3D Radar Sensing}

Point-based deep learning methods have emerged as powerful tools for 3D perception, bypassing the need for rigid voxelization. PointNet~\cite{qi2017pointnet} introduced a foundational framework that applies a symmetric function to unordered point sets, ensuring permutation invariance. It was soon extended by PointNet++~\cite{qi2017pointnet++}, which employs a hierarchical strategy to capture multi-scale neighborhood structures—an advantage for sparse radar point clouds. Subsequent models, such as DGCNN~\cite{wang2019dynamic}, leverage graph-based connectivity to dynamically refine local features, while attention-based methods integrate global context more effectively—key to handling cluttered radar returns.

Despite these advances, radar-specific challenges remain. Compared to LiDAR, radar data is sparser, noisier, and includes Doppler or cross-section values. Point-based networks must be adapted to handle high uncertainty, often via specialized sampling or clutter filtering. Exploiting Doppler as an additional feature dimension helps distinguish moving objects from background clutter, underscoring the flexibility of these architectures in capturing radar’s unique signal properties.

However, large-scale radar data demand efficient downsampling or parallelization, and adverse environments require robust training protocols for generalization. Cooperative sensing across multiple radars can further enhance perception, though issues like sensor synchronization and data association persist~\cite{xiang2023multi}. As these point-based approaches evolve to tackle such challenges, they remain central to next-generation radar perception systems.

\subsection{Cooperative Perception Dataset}



Autonomous driving datasets like KITTI~\cite{geiger2012we}, nuScenes~\cite{caesar2020nuscenes}, and Waymo Open~\cite{sun2020scalability} have been pivotal for vehicle perception, but they primarily focus on single-vehicle scenarios and do not fully address challenges stemming from heavy occlusion and long-range perception \cite{wang2020v2vnet}. Meanwhile, 2D-focused datasets such as Cityscapes~\cite{cordts2016cityscapes}, BDD100k~\cite{yu2020bdd100k}, and SYNTHIA~\cite{ros2016synthia} lack the spatial precision critical for robust navigation in complex environments.

To overcome these gaps, we introduce \textbf{CoVeRaP}, a large-scale dataset for V2V cooperative perception using FMCW radar data. Unlike LiDAR-centric datasets, CoVeRaP employs radar point clouds for more efficient real-time communication. It provides annotated 3D vehicle bounding boxes, synchronized radar point clouds, GPS data, and camera images, offering a comprehensive platform to advance cooperative perception in realistic driving scenarios.

\section{Dataset Acquisition and Preprocessing}
\label{sec:dataset}
Our experiments utilized a cooperative radar perception dataset collected using a multi-vehicle sensor platform, processed to overcome temporal and spatial synchronization challenges. The data set includes well-annotated 3D bounding-box ground truths, enabling a comprehensive evaluation of cooperative perception models.

\subsection{Sensor Platform and Real-World Data Collection}
\subsubsection{Platform setup}
\begin{figure}
\centering
    \begin{subfigure}{0.49\columnwidth}
        \centering
        \includegraphics[width=\linewidth]{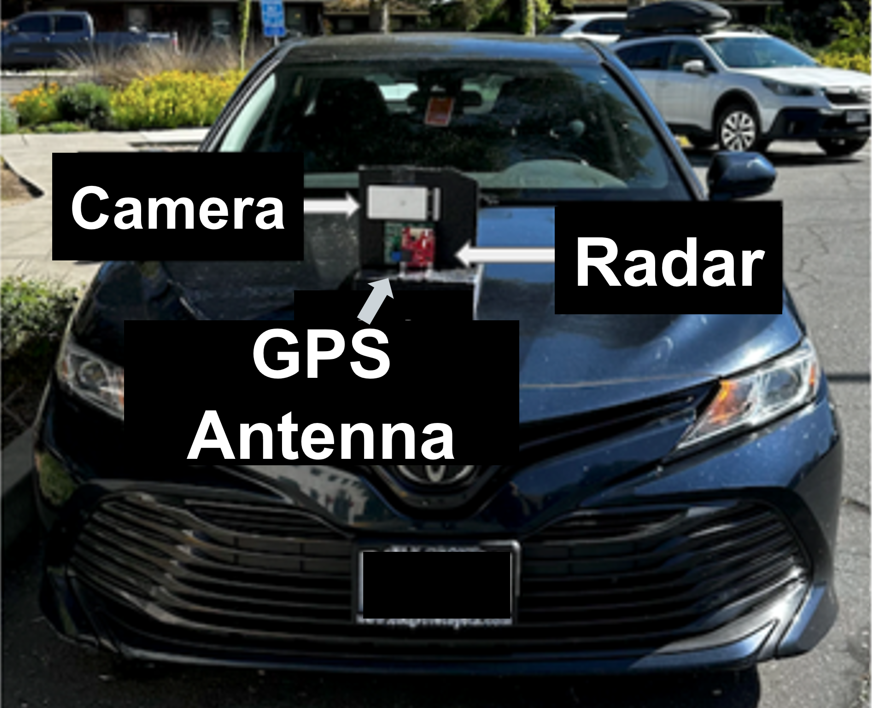}
        \caption{Ego vehicle}
        \label{fig:cars}
    \end{subfigure}
    \begin{subfigure}{0.46\columnwidth}
        \centering
        \includegraphics[width=\linewidth]{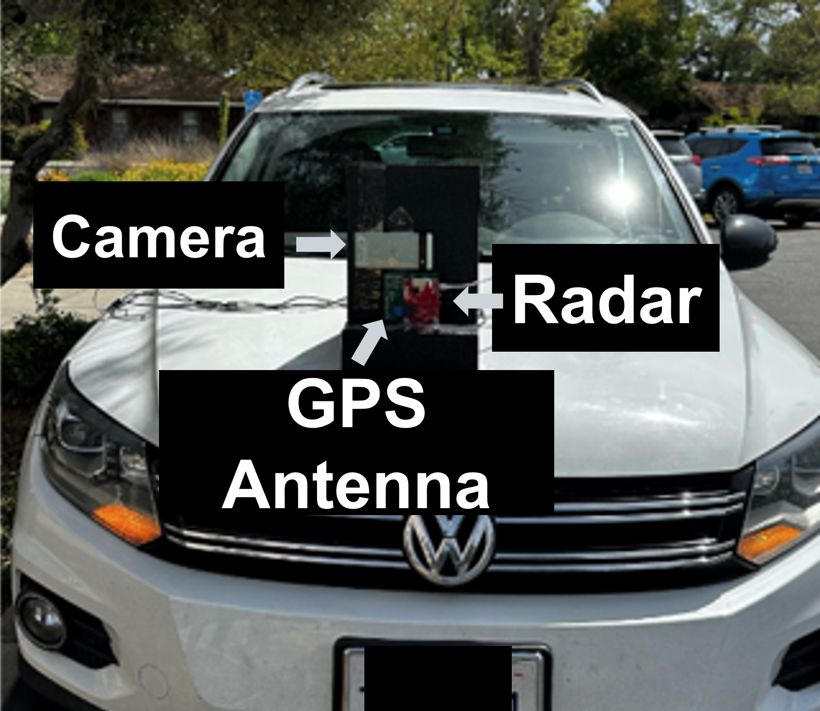}
        \caption{Assistant vehicle}
        \label{fig:sensors}
    \end{subfigure}

    \vspace{1em}

    \begin{subfigure}{0.8\columnwidth}
        \centering
        \includegraphics[width=\linewidth]{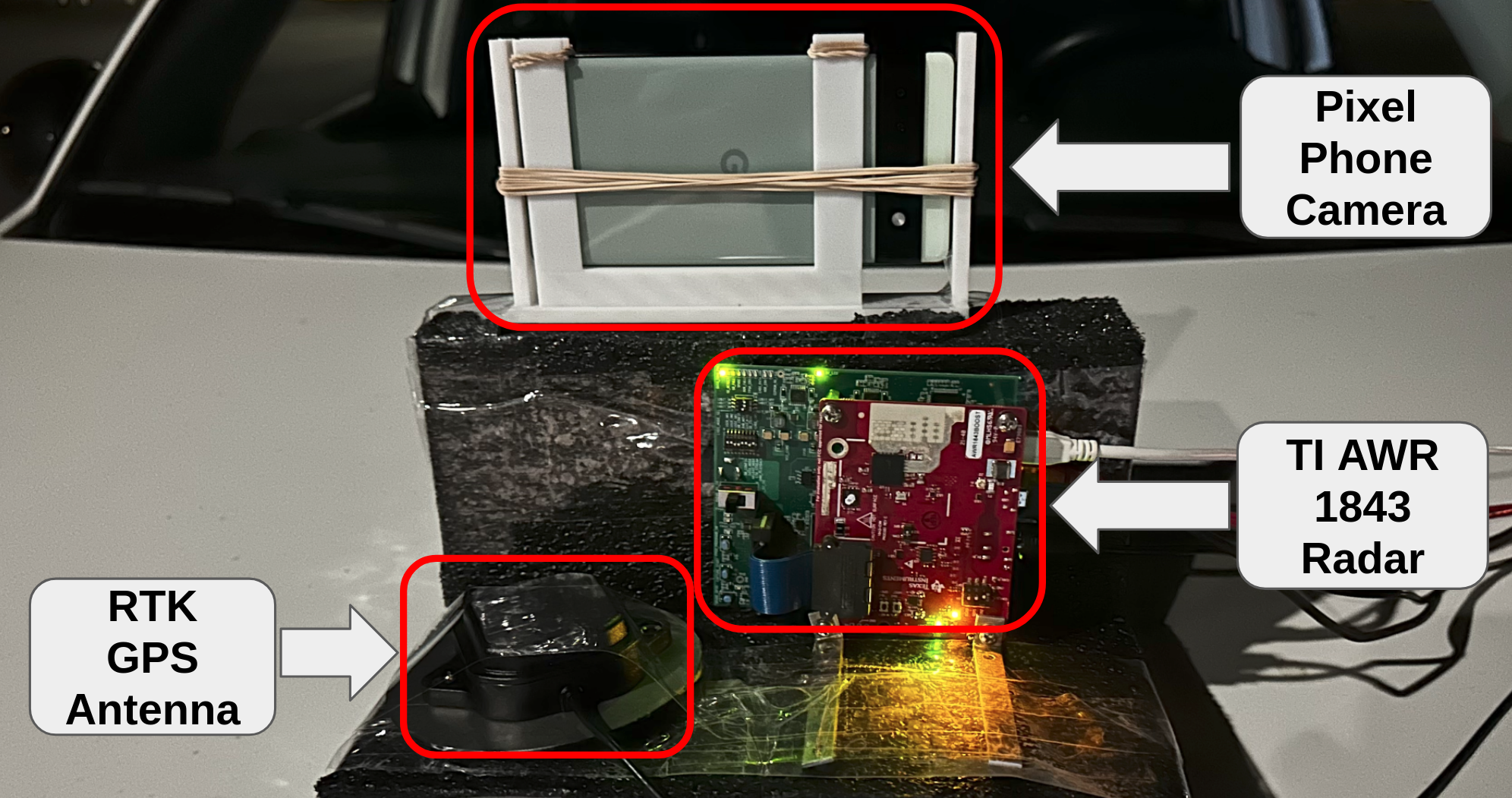}
        \caption{Sensor platform setup for sensors}
        \label{fig:third}
    \end{subfigure}

\caption{Overview of the cooperative radar perception setup. (a) Ego vehicle with FMCW radar, GPS, and camera. (b) The assistant vehicle providing complementary view radar perception. (c) Sensor alignment of radar, camera, and GPS units for synchronized data collection.}
\label{sensorsetup}
\end{figure}

Figure \ref{sensorsetup} illustrates the sensor setup used in constructing the CoVeRaP dataset. Two cooperative vehicles were deployed, each equipped with the same sensor platform and a dedicated laptop. The platform integrates an RGB camera, a GPS-RTK module, and an FMCW radar. One vehicle served as the ego view, and the other as the assistant, allowing them to exchange and transform their perception data.

\textbf{RGB Camera:} A Google Pixel 6a smartphone was used to capture timestamped image framesthat serve as ground truth for 3D bounding box annotation and ensure precise temporal and spatial alignment between sensors.

\textbf{GPS-RTK System:} We used a Sparkfun NEO-M8P-2 GPS-RTK unit paired with a GNSS multiband antenna to capture signals from both the traditional L1 and newer L2 GPS bands, achieving centimeter-level accuracy. A mobile hotspot provided real-time correction signals. Priced at under \$400, this system offers a cost-effective yet accurate alternative to conventional GPS sensors.

\textbf{FMCW Radar:} We employed the TI AWR1843 BOOST for radar sensing. Its configuration parameters are detailed in Table \ref{table:fmcw_config2}. ADC data from the radar were streamed to the laptop via a DCA1000 EVM capture board connected over Ethernet.

\textbf{Sensor Alignment and Mounting:}
For accurate calibration, the camera lens was mounted vertically above the radar antennas, and the GPS antenna was placed adjacent to the radar. Moreover, both sensor platforms were mounted at the same height on each vehicle to ensure consistency.




\begin{table}[t]
\centering
\caption{Revised FMCW Radar System Parameters}
\label{table:fmcw_config2}
\begin{tabular}{lcc}
\hline
\textbf{Parameter}             & \textbf{Value} & \textbf{Units} \\
\hline
Center Frequency ($f_{c}$)     & 77             & GHz          \\
Chirp Sweep Rate ($S$)         & 30             & MHz/$\mu$s   \\
Samples per Chirp              & 256            & --           \\
Frames Captured                & 60             & --           \\
Duration per Frame             & 100            & ms           \\
Tx, Rx Configuration           & 3, 4           & --           \\
ADC Sampling Rate              & 10,000         & ksps         \\
Azimuth Resolution             & 15             & degree       \\
Elevation Resolution           & 60             & degree       \\
\hline
\end{tabular}
\end{table}

\subsubsection{Data acquisition and synchronization}

\begin{figure}[htbp]
    \centering
    \includegraphics[width=0.8\linewidth]{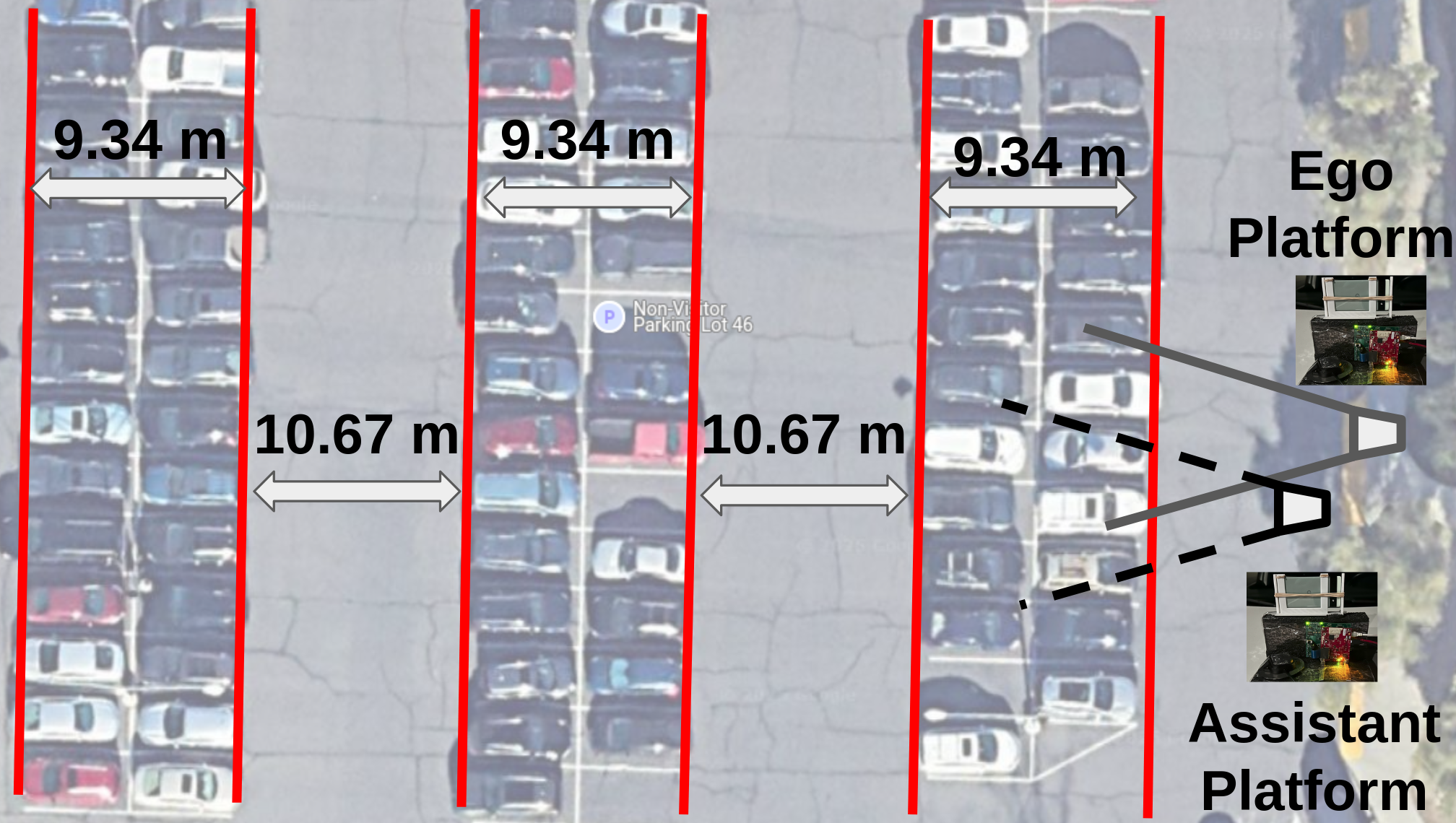}
    \caption{Illustration of the controlled parking lot environment used for dataset collection}
    \label{fig:parking_lot}
\end{figure}

Data were collected in a controlled field environment—a campus parking lot (Fig. \ref{fig:parking_lot}) with well-defined dimensions (each parking lane features two columns, each 9.34 m wide, and an inter-lane gap of 10.67 m). This consistent spatial layout, combined with a pre-defined driving route for the target vehicle, enabled precise labeling of vehicle movement. All sensors were initially synchronized using a Unix timestamp, and local video and GPS logs were recorded to support the overall data collection strategy.

\textbf{Data Acquisition Process:} During each experimental run, the target vehicle followed the pre-defined movement: forward and then backward. Two sensor platforms—one towards at the target vehicle's rear and the other along the side—continuously recorded data. The controlled campus parking lot provided reliable spatial references, ensuring that target movement between parking spots was accurately captured.

\textbf{Sensor Synchronization:} Due to varying capture frequencies and inherent timestamp shifts among sensors, we implemented a two-tier synchronization strategy consisting of intra-vehicle and inter-vehicle synchronization. 
\textit{Intra-Vehicle Synchronization:} Within each vehicle, radar data were sampled at 10 Hz. For each radar frame, we selected the camera image and the GPS coordinate with the closest timestamp,  thereby achieving temporal alignment of these sensor streams. Because the GPS operated at only 1 Hz—the lowest capture rate—we interpolated its data to 1000 Hz to match the radar's higher temporal resolution. 
\textit{Inter-Vehicle Synchronization:} To align data between the ego and assistant vehicles, we compared their initial radar timestamps. In static scenarios, where the radar mainly captured stationary point clouds, precise synchronization was less critical. However, in dynamic scenarios, even a half-second misalignment due to asynchronous delays can lead to meter-level spatial mismatches between the radar and camera data. To mitigate these discrepancies, we adopted the approach from the V2V4Real benchmark dataset~\cite{xu2023v2v4real} and selected only measurement samples with a delay of less than 100 ms for the initial alignment.

\textbf{Post-Processing and Final Data Alignment:} Residual temporal and spatial misalignments were corrected using an event-triggered synchronization method. Specifically, a sudden increase in radar velocity with strong signal strength—indicating the onset of motion—was used to recalibrate sensor timestamps. We also visualized camera frames to verify these motion timestamps, which allowed us to accurately compute the timestamp drift between radar and camera data. Notably, within a given scene, the drift remains consistent across sensors; that is, although sensors may have different initial timestamps, they share the same motion period. Using this observation, we aligned all radar, camera, and GPS frames accordingly. Concurrently, the interpolated GPS data were used to calculate spatial offsets between the ego and assistant vehicles, ensuring that multi-vehicle radar point clouds were accurately aligned and transformed between the respective vehicle coordinate systems.

In total, we captured 100 scenarios, and each scenario consists of 600 frames with a duration of 100 ms per frame, resulting in a 60-second recording per scenario.

\subsection{Dataset Construction}
After raw sensor data collected from our cooperative vehicles and its pre-processing and alignment, this section discusses the full dataset construction for cooperative perception tasks. Specifically, we generated four datasets in three types:

\subsubsection{Overview \& Components}
\begin{itemize} 
\item \textbf{Rear-View Dataset:} Sensor data captured exclusively from the rear view of the target vehicle. 
\item \textbf{Side-View Dataset:} Sensor data obtained from one side of the target vehicle. 
\item \textbf{Fused Ego-Rear Dataset:} In this variant, the ego vehicle supplies the rear-view data, while the assistant vehicle provides the side-view data. The side-view radar data are spatially shifted—using GPS offsets—to align with the rear-view coordinate system before merging. 
\item \textbf{Fused Ego-Side Dataset:} Conversely, the rear-view data are shifted into the side-view coordinate system and merged with the side-view data. 
\end{itemize}

Each of these datasets includes synchronized radar frames, camera frames, GPS frames, and ground truth annotations that follow a common index sequence. Event-triggered synchronization ensures that frames are consistently ordered across all sensors. Specifically, each radar CSV file contains fundamental features such as X, Y, and Z coordinates, range, velocity, bearing, and intensity. The camera frames provide a timestamp and an RGB image to capture motion events and support ground truth annotation, while the GPS data offer earth-centered coordinates along with timestamps for calculating the relative distances between sensor platforms.

\subsubsection{Point cloud processing and filtering}

We generated 3D point clouds by adapting an open-source codebase~\cite{xue2021mmmesh} to process our raw sensor measurements. Rather than applying a fixed threshold via CA-CFAR—which can lead to significant fluctuations in the number of detected points between frames—we dynamically selected candidate points by choosing the top 128 highest-value points from the Doppler-FFT heatmap for each frame. This method stabilizes point cloud density across frames, which is essential in dynamic and occluded environments where energy distribution may vary.

To improve reliability, we further filtered the candidate points by discarding those likely affected by multipath effects—reflections that typically exhibit lower energy than direct signals—and by removing points that fell outside a plausible distance range from the radar. Finally, we retained only the top 20\% of points based on signal intensity to ensure the quality and consistency of the resulting 3D point clouds.

\begin{figure}[t]
\centering
    \begin{subfigure}{\columnwidth}
        \centering
        \includegraphics[width=0.5\columnwidth]{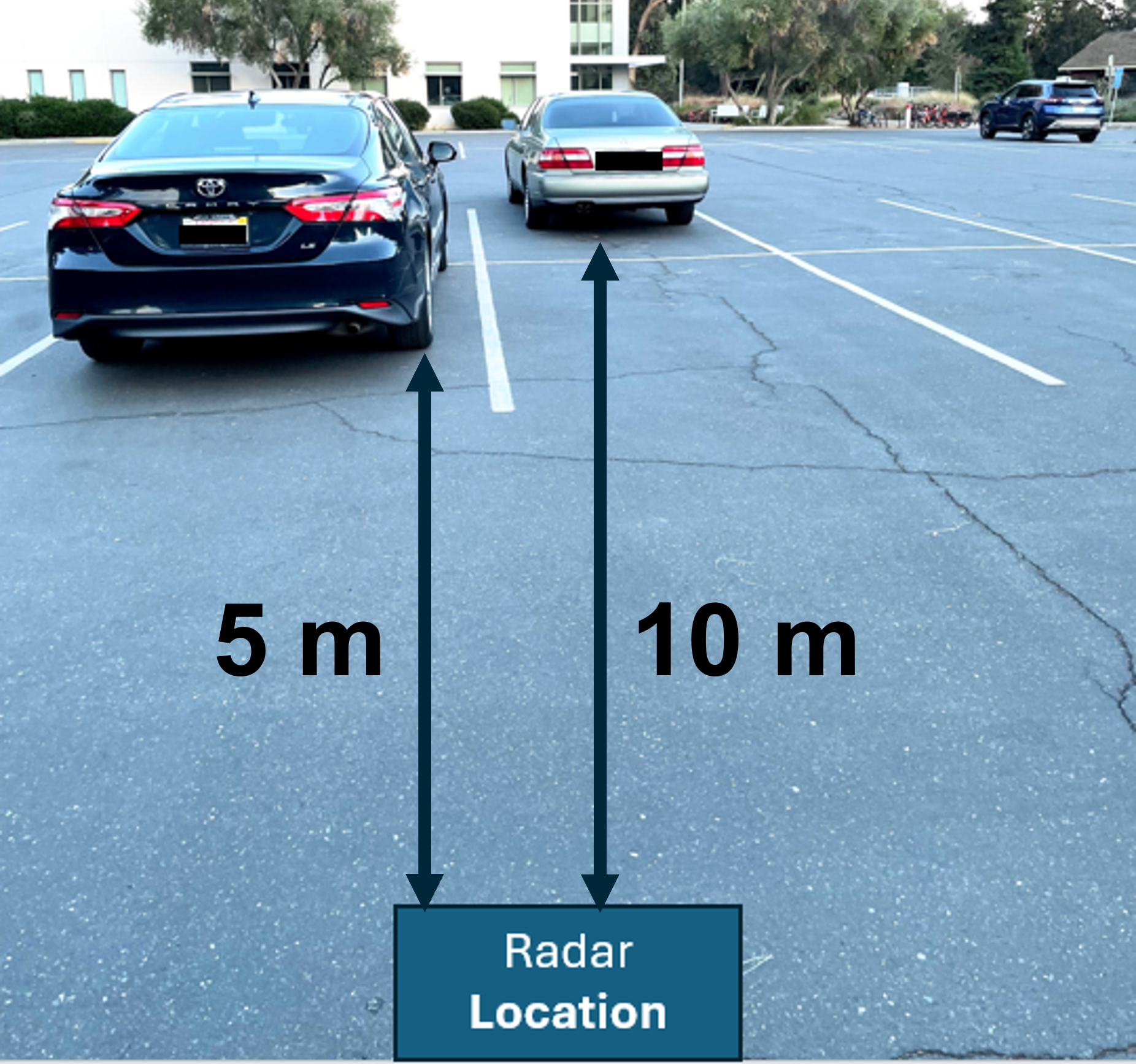}
        \caption{RGB camera frame from the rear-view}
        \label{fig:PCF1}
    \end{subfigure}
    
    \begin{subfigure}{\columnwidth}
        \centering
        \includegraphics[width=1\columnwidth]{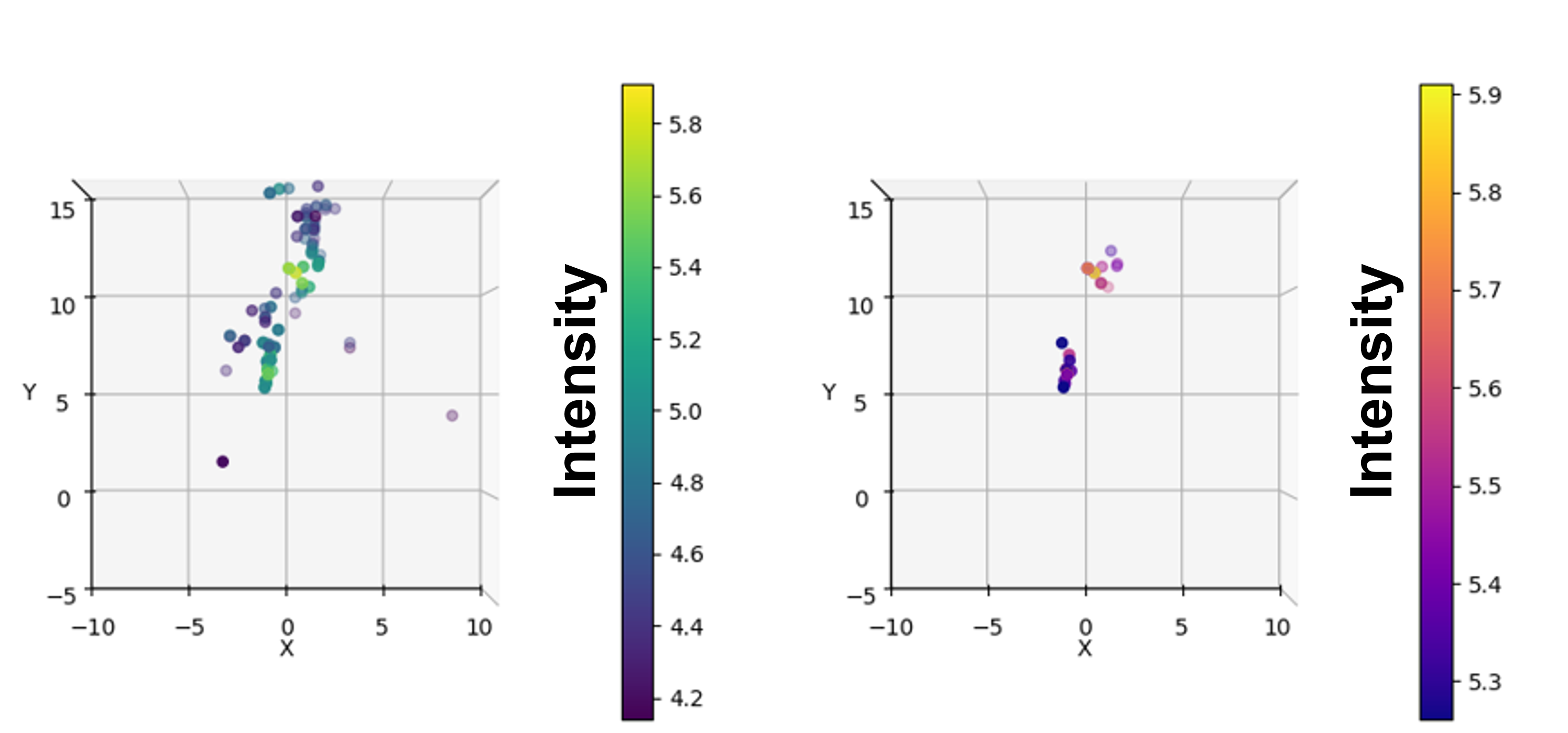}
       \caption{Bird’s-eye view of radar point clouds. Left: raw unfiltered points. Right: top 20\% points, filtered by signal intensity.}
        \label{fig:PCF2}
    \end{subfigure}
\caption{Overview of rear-view radar sensing: camera view (top) and corresponding 3D point clouds (bottom), which illustrates a raw radar detection and one filtered by signal intensity.}
\label{occul_assistant}
\end{figure}

Figure \ref{fig:PCF2} illustrates our processing pipeline: (a) shows a sample raw measurement on the target vehicle, (b) displays the initial 3D point cloud, and (c) presents the point cloud after applying the top 20\% intensity filtering.

\subsubsection{Development of ground truth labels}

Ground truth annotations, in the form of 3D bounding boxes, were generated by interpolating pre-defined motion parameters for each scene. Each scene was divided into distinct phases: an initial static phase, a forward motion phase, a brief pause, and a backward motion phase. During the initial static phase (e.g., frames 0 to $x$), the vehicle remains stationary. Starting from frame $x+1$, the vehicle moves forward until frame $x+f$, then pauses briefly before reversing until it comes to a final rest. For each phase, starting and stopping positions (X, Y, Z), fixed vehicle dimensions (length, height, width), orientation, and bearing were provided. These parameters were used to interpolate the 3D bounding box center for every frame, ensuring that the annotations accurately reflect the vehicle’s motion and size.

To further ensure the accuracy of our annotations, we validated the sensor data by matching radar frames with their corresponding camera frames. We visualized radar detection plots, radar features from the CSV files, and the associated camera frames containing ground truth information. Based on these visual inspections, only radar frames with verified detections were retained; frames with noise or abnormal detections were discarded. Ground truth 3D bounding box annotations were applied only to these validated frames.

\subsection{Data Splitting and Final Data Format}

To facilitate robust model evaluation, each scene was partitioned into training, validation, and test subsets by 80\%–10\%–10\% ratio. These per-scene splits were then aggregated to form unified training, validation, and test. For consistency, the same index order was maintained among the rear-view, side-view, and fused-view datasets.

The final datasets are stored as a collection of \texttt{.npy} files. Each dataset includes arrays for inputs corresponding to that view, the ground truth labels, and index arrays defining the data splits. This format ensures straightforward data access, and facilitates future research in cooperative perception.
\section{System Architecture and Model Design}
\label{sec:system_and_model}
This section describes our cooperative radar perception system, which leverages multi-vehicle sensor data to enhance 3D object detection. Our approach is built on a comprehensive model that integrates multi-branch feature extraction, attention-based fusion, and sequence-aware processing. In what follows, we first present the overall system architecture and the applied fusion strategies, then detail the baseline model architecture, and finally provide key implementation details.

\subsection{Overall System Architecture in Fusion Strategies} 
\begin{figure}[htbp]
    \centering
    \includegraphics[width=1.0\linewidth]{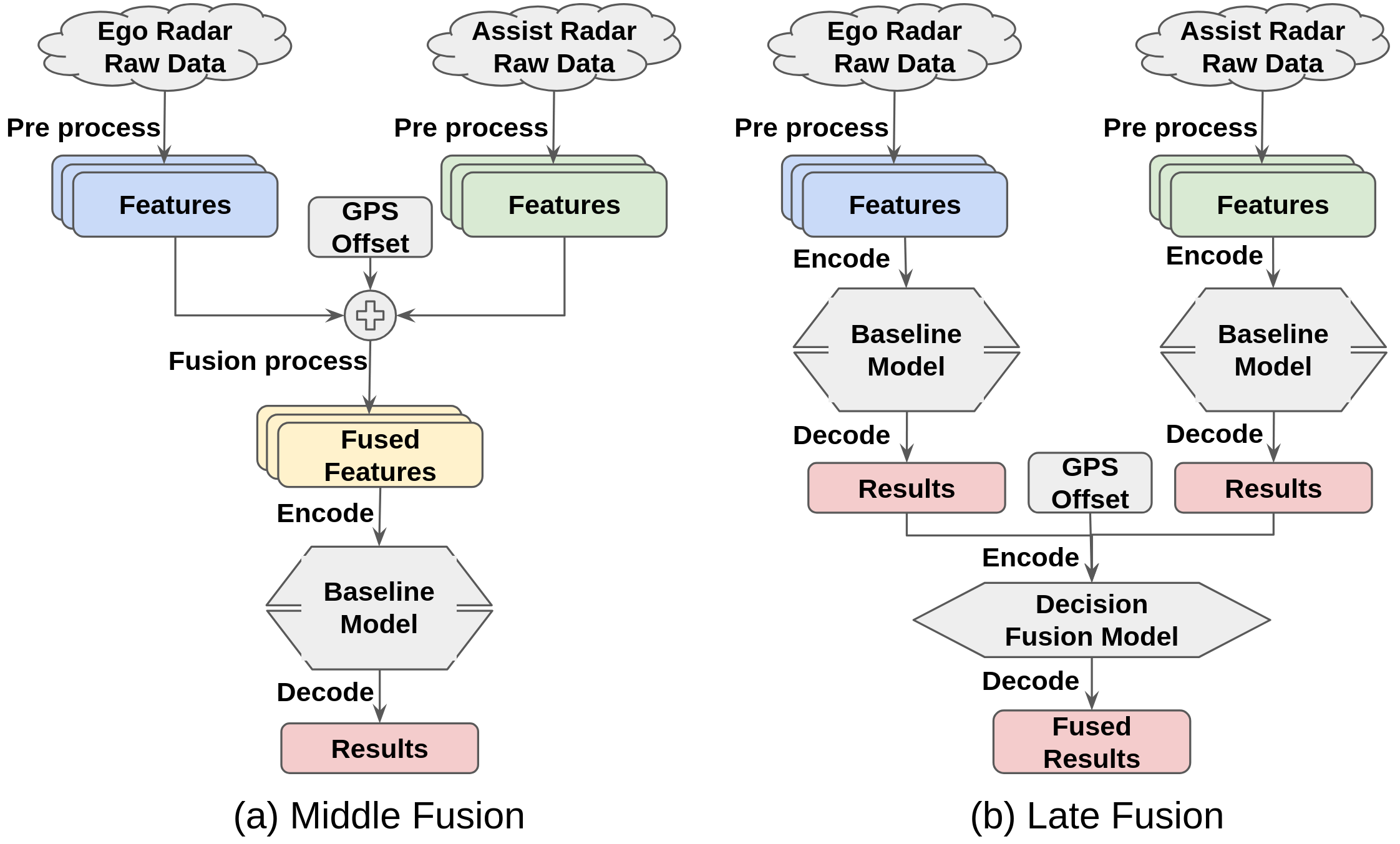}
    \caption{Two  fusion strategies for multi-vehicle radar perception. (a) Middle Fusion: Feature-level integration of radar data from multiple vehicles in a common coordinate frame before object detection. (b) Late Fusion: Independent prediction at each vehicle, followed by prediction-level fusion of detected bounding boxes.}
    \label{fig:two_fusions}
\end{figure}

In our cooperative radar perception, two principal fusion strategies are adopted: \emph{middle fusion} and \emph{late fusion} as shown in Fig.~\ref{fig:two_fusions}. Both strategies ultimately aim to improve the 3D object detection performance by leveraging complementary sensor data from multiple vehicles; however, they differ in how and when the data are integrated.

\textbf{Middle Fusion:}
In the middle fusion approach, raw sensor data are collected from each vehicle, and then data features are extracted before the data is synchronized among vehicles and aligned into a common coordinate frame. This fusion occurs at the feature level prior to any final 3D boundng box prediction. Specifically, the sequential stages in our middle fusion pipeline are as follows: 
\begin{enumerate} 
\item \textbf{Multi-Vehicle Sensing:} Each vehicle independently collects radar point clouds, along with supporting GPS and camera data, which is a ground truth for label annotation. 
\item \textbf{Time and Spatial Alignment in Data Fusion:} This synchronization procedure not only ensures temporal consistency but also minimizes spatial calibration errors, which is critical for reliable fusion in dynamic driving scenarios. The event-trigger mechanism is employed to synchronize the sensor data from all radars and cameras in time and space, while interpolated GPS offsets are used to shift the other radar data into ego vehicle's sensing coordinate system. Then, the synchronized data are fused for the ego vehicle.  
\item \textbf{Enhanced ego perception:} The fused multiple sensing data are fed into our baseline deep learning model, which has three branches (processing position, dynamics, and intensity)to learn features and a self-attention module that integrates the outputs from the three branches. After processing the fused features, the model predicts 3D bounding boxes for the target vehicle. 
\end{enumerate}
 
\textbf{Late Fusion:}
Alternatively, in the late fusion, each vehicle operates independently throughout the feature-learning and prediction stages. No inter-vehicle data enhancement occurs during model training. Instead, each vehicle’s radar model is trained solely on the local sensor data of that vehicle, by same baseline architecture. The process then proceeds as follows:
\begin{enumerate} 
\item \textbf{Individual Sensing and Learning:} Each vehicle collects and processes its own radar data. The baseline model is trained locally to detect and regress 3D bounding boxes from that vantage point, without sharing intermediate features or raw data. 
\item \textbf{Independent Prediction:} After training, each vehicle’s model independently infers bounding boxes and confidence scores in its local coordinate frame.
\item \textbf{Coordinate Alignment:} To integrate these predictions into a common reference system, we apply rigid transformations (GPS-offsets) that map the peer-vehicle bounding boxes into the ego vehicle’s coordinate frame.
\item \textbf{Decision-Level Fusion:} Finally, a decision layer which is a MLP neural network combines bounding boxes based on their corresponding predicted confidence scores. The end result is a single fused 3D bounding box representation for the detected vehicle, reflecting information from both vantage sensing results.
\end{enumerate}

Late fusion thus encapsulates each vehicle’s detection pipeline as a separate module, preserving autonomy and minimizing data transmission (only bounding boxes and confidence scores). This design also allows the same baseline model architecture to be reused on each vehicle independently, while still providing a mechanism to combine final predictions.

\subsection{Baseline Model Architecture} 
\begin{figure*}[t]
  \centering
  \includegraphics[width=\textwidth]{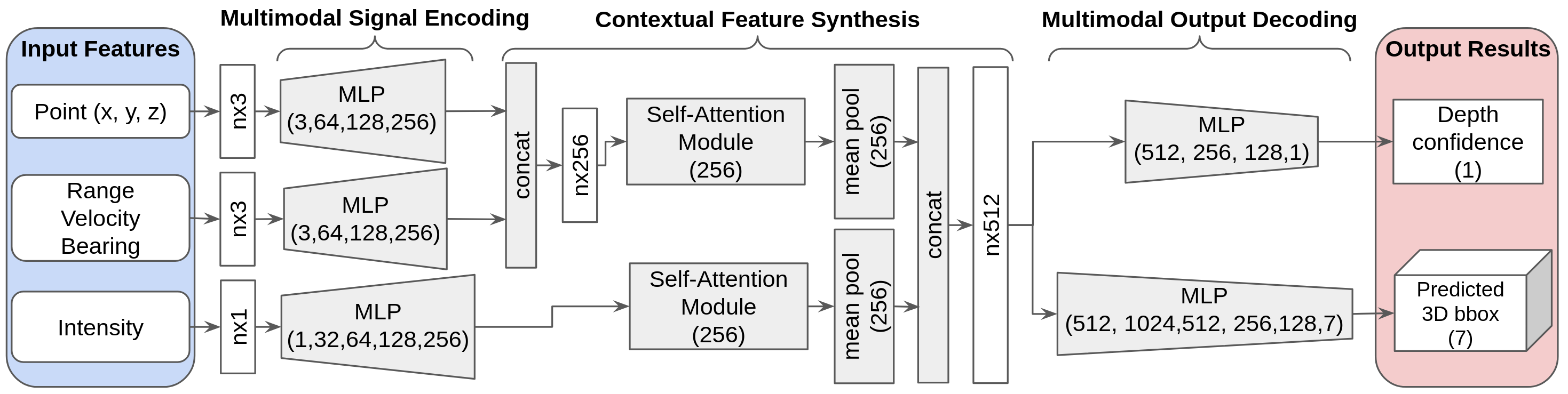}
  \caption{Detailed architecture of the proposed cooperative radar perception model. The model consists of three feature extraction branches (spatial position, dynamics, and intensity), followed by an attention-based fusion module and prediction heads for depth estimation and 3D bounding box regression.}
  \label{fig:baseline_model}
\end{figure*}

To address the challenges of noisy, sparse radar data, our baseline model (see Fig.~\ref{fig:baseline_model}) employs a multi-branch architecture that integrates feature extraction with attention mechanisms, inspired by PointNet and self-attention.

The proposed model follows a three-stage pipeline: (1) a multimodal signal encoding stage, which extracts spatial, dynamic, and intensity-based features from radar point clouds; (2) a contextual feature synthesis stage, which integrates these features through concatenation, self-attention, and pooling; and (3) a multimodal output decoding stage, which transforms the fused representation into 3D bounding box predictions and depth confidence scores.

\subsection{Multimodal Signal Encoding}

The encoding stage decomposes radar input into three distinct feature streams: position, dynamics, and intensity. The position branch processes 3D spatial coordinates $(x,y,z)$ using a series of fully connected multi-layer perceptron (MLP) layers, each followed by Layer Normalization, ReLU activation, and Dropout. Inspired by PointNet, this design progressively expands the feature representation from 3 to 256 dimensions, allowing the model to retain spatial structure while performing per-point feature extraction without losing local neighborhood relationships. The dynamics branch encodes motion-related features—including velocity, range, and bearing—into a 256-dimensional latent space. Since these features are inherently correlated, this branch enables the model to isolate moving objects from static background clutter, thereby improving object localization and tracking in noisy radar environments. The intensity branch is responsible for capturing the 1D intensity values of radar reflections. A small MLP first expands the intensity dimension from 1 to 256, followed by a learned intensity attention mechanism that assigns higher weights to points with stronger radar returns. Empirically, valid detections exhibit significantly higher intensity than background noise, and leveraging this observation enhances robustness against spurious detections and noisy reflections.

\subsection{Contextual Feature Synthesis}

Once the raw radar features are extracted, they are integrated into a unified representation to preserve complementary information across spatial, motion, and intensity-based cues. The outputs of the position and dynamics branches, each with 256 dimensions, are concatenated into a 512-dimensional feature vector and projected back to 256 dimensions through a linear fusion layer. A self-attention module is applied to refine these point-wise features, followed by mean pooling to obtain a single 256-dimensional global context vector. Meanwhile, the intensity branch generates an independent 256-dimensional feature map, which undergoes intensity-aware pooling to filter out low-confidence radar returns. The final fused representation is obtained by concatenating these two feature vectors, forming a 512-dimensional embedding that encodes geometric structure, motion characteristics, and signal intensity within a compact latent space.

\subsection{Multimodal Output Decoding}

The final decoding stage translates the fused representation into actionable predictions using two dedicated MLPs. The depth estimation subnet predicts a confidence score for depth estimation based on radar signal characteristics. Many existing depth estimation models \cite{piccinelli2024unidepth, wu2019detectron2, yang2024depth} struggle to generalize across domains, leading to large errors in out-of-distribution scenarios. To address this, the proposed model includes a depth confidence estimator tailored specifically to radar data, ensuring robust and reliable depth predictions that directly impact 3D bounding box regression. The 3D bounding box decoder further processes the fused embedding by expanding the feature representation to 1024 dimensions, increasing the model’s capacity to capture nonlinear interactions in the fused feature space. The final output consists of seven key parameters $(w,h,l,x,y,z,\theta)$, representing the object's width, height, length, center coordinates, and orientation.
\section{Experiment \& Results}
\label{sec:experiment_and_results}

\subsection{Implementation Details} 
\textbf{Dataset:}
In the evaluation, we select 11 high-quality scenes (21,116 frames) that pass strict synchronization and annotation checks and together span diverse radar-view geometries. These scenes include (i) rear+side complementary views, where one radar mostly sees the back of the target while the other views the flank, and (ii) partial-overlap configurations in which both radars share part of the field of view, creating denser but high-clutter point clouds. Such combinations stress the detector with sparser returns and Doppler clutter, providing a realistic yet controlled benchmark. Each scene is split into training, validation, and test sets in an 80\%/10\%/10\% ratio, resulting in 16,874 training frames, 2,116 validation frames, and 2,126 test frames overall. This division ensures that our model learns from diverse scenarios and is robustly evaluated. For each frame, the top 70 points—selected based on signal intensity—are retained, capturing both valid reflections and noise, while each frame is precisely aligned with a corresponding GPS offset for accurate spatial calibration. Furthermore, each scene includes two radar sensing directions, namely the rear view (located at the back of the target vehicle) and the side view (located at the side), providing complementary perspectives for cooperative sensing.

\textbf{Middle Fusion:}
In our middle-fusion strategy, radar data from multiple vehicles is integrated at the feature level using a common coordinate system — ego vehicle first, then assistant. First, each vehicle’s radar point cloud (comprising spatial coordinates, dynamics, and intensity) is aligned by applying GPS offset adjustments. For instance, we first treat the rear view as the ego and shift the side view data into ego's coordinate frame; then, we reverse the roles, treating the side view as the ego and transform the rear view data. The aligned data streams are concatenated into a unified feature stack and fed into model’s multimodal signal encoding stage. This feature sharing across vehicles enables the network to learn complementary representations from other radar returns, ultimately improving detection robustness.

\textbf{Late Fusion:}
In the late-fusion approach, each vehicle independently predicts a 3D bounding box (7 dimensions) and a confidence score (1 dimension) in its local frame. Non-ego detections are transformed into the ego frame using GPS offsets. The outputs are concatenated into a 16-dimensional vector and passed through a lightweight fusion MLP that expands to 128 dimensions, then reduces to 64, and finally outputs an 8-dimensional vector—7 dimensions for the final 3D bounding box and 1 for the fused confidence score. This method produces robust final detections through confidence-weighted fusion.

\textbf{Training \& Evaluation:}
Our model is implemented in PyTorch and trained for 600 epochs using a batch size of 8 and a learning rate of 0.0005—hyperparameters tuned via validation. We employ the AdamW optimizer with a weight decay of 0.001 and a ReduceLROnPlateau scheduler to ensure stable convergence, and all experiments are conducted on an NVIDIA RTX A6000 GPU. Both our middle fusion and late fusion approaches use the same data splits and a composite loss function that combines Smooth L1 loss for bounding-box regression, binary cross-entropy for depth confidence, and a custom IoU loss that penalizes misaligned predictions. Model performance is continuously monitored on a separate test set, and checkpoints with the best validation loss are saved for final evaluation on the test set, where loss values and 3D IoU metrics are reported.


\subsection{Dataset Evaluation \& Analysis}

\begin{figure}[t]
  \centering
  \includegraphics[width=0.9\columnwidth]{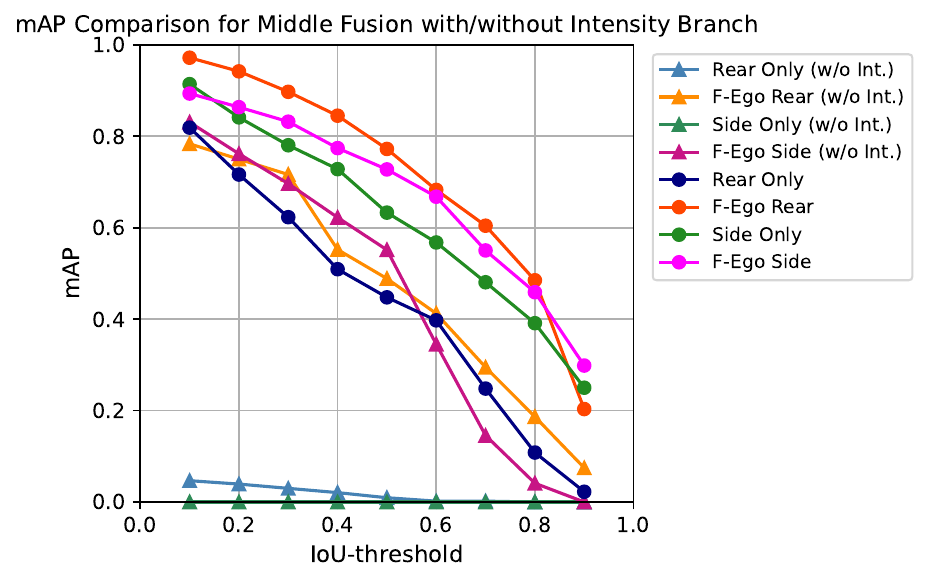}
  \caption{Impact of radar-return intensity \textbf{(Int.)} encoding on Middle-Fusion performance across IoU thresholds for both single-view and \textbf{Fused-Ego (F-Ego)} settings. Curves with circle markers include Int. features and show a clear mAP lift—especially at stricter IoU levels—while the F-Ego curves demonstrate the additional gain obtained by fusing complementary views.}
  \label{fig:results_MF_w_wo_intensity}
\end{figure}

\begin{figure}[t]
  \centering
  \includegraphics[width=0.9\columnwidth]{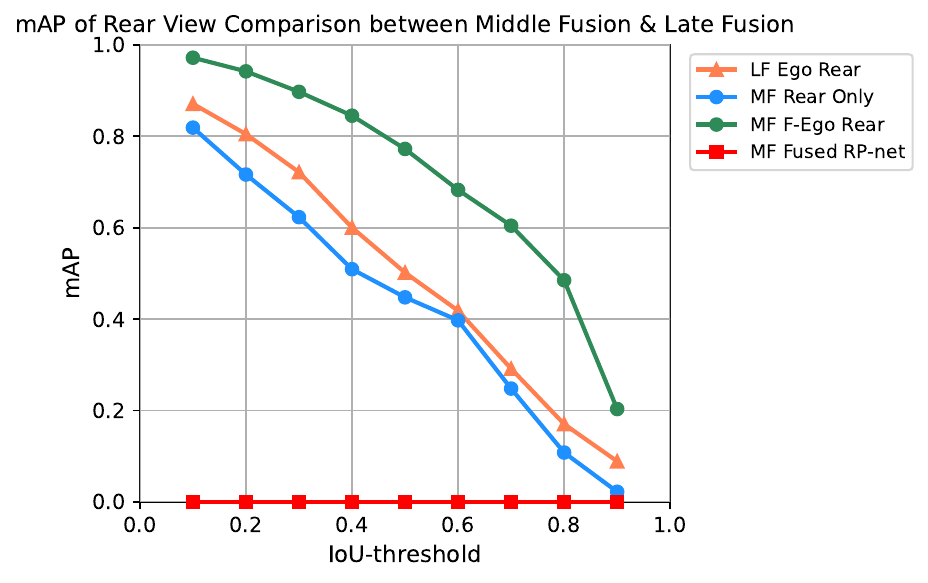}
  \caption{Comparison of Middle Fusion and Late Fusion approaches for 3D object detection using cooperative radar perception. Middle Fusion (MF) consistently outperforms Late Fusion (LF) across all IoU thresholds, demonstrating the advantages of feature-level integration over decision-level fusion.}
  \label{fig:results_MF_LF}
\end{figure}

Figure \ref{fig:results_MF_w_wo_intensity} presents the performance of Middle Fusion under eight configurations: rear view and side view, each evaluated in two settings—using the baseline model with intensity encoding (Int.; radar-return intensity) and without intensity encoding—and with both singular and fused views. The plots, measured across IoU thresholds from 0.1 to 0.9, clearly show that incorporating intensity in the feature encoding stage (represented by the circle markers) consistently boosts mAP, particularly at stricter thresholds. For example, the "Side Only (w/o Int.)" configuration fails to yield any correct predictions (0 mAP) across all thresholds, while the "Side Only" configuration with intensity encoding achieves above 0.91 mAP at an IoU threshold of 0.1 and 0.25 mAP at 0.9. This improvement highlights the critical role of intensity in distinguishing valid radar returns from noise.

Additionally, this figure demonstrates that the fused-ego (F-Ego) view consistently outperforms the corresponding singular view. Specifically, the "F-Ego Rear" curve (displayed in orange) remains above the "Rear Only" curve (shown in navy), and the "F-Ego Side" curve (in magenta) nearly overlaps or exceeds the "Side Only" curve (in forest green) across all IoU thresholds. This indicates that fusing data from multiple perspectives further enhances detection performance by leveraging complementary information from different views.   

Figure \ref{fig:results_MF_LF} presents a comparison of rear-view performance between Middle Fusion and Late Fusion strategies. In this plot, “MF Rear Only” and “MF Fused Ego Rear” denote two middle fusion variants—one using only the rear view and the other fusing features from the complementary view—while “LF Ego Rear” represents the late fusion approach, in which the ego (rear) view is enhanced with predictions from the other view. Notably, the MF Fused Ego Rear configuration consistently outperforms the others across all IoU thresholds.
Additionally, the competitor model RP-net~\cite{bansal2020pointillism} achieves 0 mAP across all thresholds within our middle fusion framework. This failure is attributed to its inability to perform robust feature fusion under changing relative radar positions and its lack of an effective mechanism to handle noisy points—capabilities that our intensity encoding module successfully provides.

Table \ref{tab:middle_fusion_performance} and Table \ref{tab:fusion_comparison} provide the performance numbers in tabular form for quick reference, which list exact mAP values and ratios of outperforming. Table \ref{tab:middle_fusion_performance} covers middle fusion with intensity for both rear and side views, including ratios that indicate how many times a fused view is better than a single view. Table \ref{tab:fusion_comparison} specifically addresses rear-view performance across various IoU thresholds for two fusion strategies, with or without intensity.

Overall, these results confirm that fusing radar data—whether via middle fusion or late fusion—improves detection performance over single-view baselines. Middle fusion, however, tends to outperform late fusion, especially at higher IoU thresholds (e.g., 0.7–0.9), because feature-level integration retains more information than simple bounding-box merging. Additionally, intensity encoding in the middle fusion architecture significantly boosts accuracy, particularly in sparse or noisy radar conditions. The tabulated ratios highlight that combining ego and assistant views (e.g., “Fused Ego Rear” vs. “Rear Only”) can yield improvements of up to 9 times at stricter thresholds (0.9). These consistent gains underscore the importance of sensor-level data fusion and intensity-based attention for robust radar perception.

\begin{table*}[t]
  \centering
  \caption{Performance of Middle Fusion (with intensity) across IoU thresholds}
  \label{tab:middle_fusion_performance}
  \begin{tabular}{lccccccc}
    \toprule
    \textbf{IoU} & 
    \shortstack{\textbf{}\\\textbf{Rear Only}} & 
    \shortstack{\textbf{}\\\textbf{Fused Ego Rear}} & 
    \shortstack{\textbf{}\\\textbf{Side Only}} & 
    \shortstack{\textbf{}\\\textbf{Fused Ego Side}} & 
    \shortstack{\textbf{Fused Ego Rear}\\\textbf{/ Rear}} & 
    \shortstack{\textbf{Fused Ego Rear}\\\textbf{/ Fused Ego Side}} & 
    \shortstack{\textbf{Fused Ego Side}\\\textbf{/ Side}} \\
    \midrule
    0.9 & 0.0224 & 0.2034 & 0.2500 & 0.2985 & 9.0833 & 0.6813 & 1.1940 \\
    0.8 & 0.1082 & 0.4851 & 0.3912 & 0.4590 & 4.4828 & 1.0569 & 1.1733 \\
    0.7 & 0.2481 & 0.6045 & 0.4807 & 0.5504 & 2.4361 & 1.0983 & 1.1449 \\
    0.6 & 0.3974 & 0.6828 & 0.5678 & 0.6679 & 1.7183 & 1.0223 & 1.1763 \\
    0.5 & 0.4478 & 0.7724 & 0.6331 & 0.7276 & 1.7250 & 1.0615 & 1.1493 \\
    0.4 & 0.5093 & 0.8451 & 0.7282 & 0.7743 & 1.6593 & 1.0916 & 1.0632 \\
    0.3 & 0.6231 & 0.8974 & 0.7805 & 0.8321 & 1.4401 & 1.0785 & 1.0661 \\
    0.2 & 0.7164 & 0.9422 & 0.8414 & 0.8638 & 1.3151 & 1.0907 & 1.0266 \\
    0.1 & 0.8190 & 0.9720 & 0.9142 & 0.8937 & 1.1868 & 1.0877 & 0.9776 \\
    \bottomrule
  \end{tabular}
\end{table*}


\begin{table*}[t]
  \centering
  \caption{\textbf{Rear View} Performance across IoU Thresholds for Two Fusion Strategies with and without Intensity}
  \label{tab:fusion_comparison}
  \begin{tabular}{lccccc}
    \toprule
    \textbf{IoU} & 
      \shortstack{\textbf{Middle Fusion}\\\textbf{(with intensity)}} & 
      \shortstack{\textbf{Middle Fusion Fused View}\\\textbf{(with intensity)}} & 
      \shortstack{\textbf{Middle Fusion}\\\textbf{(without intensity)}} & 
      \shortstack{\textbf{Middle Fusion Fused View}\\\textbf{(without intensity)}} & 
      \shortstack{\textbf{Late Fusion}\\\textbf{(Ego Rear)}} \\
    \midrule
    0.9 & 0.0224 & 0.2034 & 0.0000 & 0.0746 & 0.0890 \\
    0.8 & 0.1082 & 0.4851 & 0.0000 & 0.1866 & 0.1705 \\
    0.7 & 0.2481 & 0.6045 & 0.0019 & 0.2948 & 0.2917 \\
    0.6 & 0.3974 & 0.6828 & 0.0019 & 0.4123 & 0.4186 \\
    0.5 & 0.4478 & 0.7724 & 0.0093 & 0.4888 & 0.5019 \\
    0.4 & 0.5093 & 0.8451 & 0.0205 & 0.5522 & 0.6004 \\
    0.3 & 0.6231 & 0.8974 & 0.0299 & 0.7164 & 0.7216 \\
    0.2 & 0.7164 & 0.9422 & 0.0392 & 0.7500 & 0.8049 \\
    0.1 & 0.8190 & 0.9720 & 0.0466 & 0.7836 & 0.8712 \\
    \bottomrule
  \end{tabular}
\end{table*}


\section{Conclusion}
\label{sec:conclusion}
We presented \textbf{CoVeRaP}, the first cooperative mmWave-radar dataset with time-aligned radar, camera, and GPS streams from multiple vehicles, creating a solid testbed for V2V sensing.
Building on CoVeRaP, we designed a unified framework that supports both middle (feature-level) and late (prediction-level) fusion. Its baseline detector uses a multi-branch, attention-enhanced backbone to merge spatial, Doppler, and intensity cues before outputting 3-D bounding boxes and depth confidence.

Experiments confirm the value of cooperation: on the fused ego + rear view, middle fusion with intensity encoding lifts mAP by 9 × at IoU 0.9 over a single rear view, and fused views always surpass individuals, with middle fusion consistently beating late fusion. Models without intensity encoding fail on the side view data, underscoring importance of this feature.

Limitations \& future work. The public release currently contains only parallel-lane scenes and vehicle targets. We have already captured—but not yet processed—lane-merge/lane-change runs and T-intersection traffic, where the two radars form a 90° “T” configuration and provide non-line-of-sight coverage. Because our platform lacks LiDAR, we intentionally explore how far a low-cost radar-only suite can be pushed. Incoming releases will add these complex trajectories and expand the object classes to bikes, pedestrians, and road signs.

The source code and collected data for reproducing our experiments and analyses are publicly available at \url{https://github.com/John1001Song/FMCW_Vehicle_Fusion}.

\section{Acknowledgment}
We would like to thank the anonymous reviewers for their valuable comments and feedback. This research is supported by ARL grant W911NF22-1-0216



\bibliographystyle{ieeetr}
\bibliography{references}

\end{document}